\documentstyle[11pt]{article}

\newcommand{\lingform}[1]{{\it #1}}

\title{Building Natural Language Generation Systems}
\author{Ehud Reiter\\
Department of Computing Science\\
University of Aberdeen\\
King's College\\
Aberdeen AB9 2UE, BRITAIN\\
email: {\tt ereiter@csd.abdn.ac.uk}}

\date{ }

\begin{document}

\pagestyle{empty}
\bibliographystyle{alpha}
\setlength{\baselineskip}{13pt}               

\maketitle

\section{Introduction}

Natural Language Generation (NLG) systems generate texts in English
(or other human
languages, such as French) from computer-accessible data.  NLG systems are
(currently) most often used to help human authors write routine documents,
including business letters \cite{springer:iaai} and
weather reports \cite{goldberg:ieee94}.  They also have been used as
interactive explanation tools which communicate information in
an understandable way to non-expert users, especially in software
engineering (eg, \cite{joyce:anlp}) and medical (eg, \cite{pitt:medical})
contexts.

{}From a technical perspective, almost all applied NLG systems perform the
following three tasks \cite{reiter:arch}:
\begin{description}
\item[Content Determination and Text Planning:]
Decide what information should be communicated to the user (content
determination), and how this information should be rhetorically structured
(text planning).  These tasks are usually done simultaneously.
\item[Sentence Planning:]
Decide how the information will be split among individual sentences and
paragraphs, and what cohesion devices
(eg, pronouns, discourse markers)
should be added to make the text flow smoothly.
\item[Realization:]
Generate the individual sentences in a grammatically correct manner.
\end{description}
In the rest of this paper, I shall briefly discuss different ways of
performing each of these tasks

\section{Content Determination and Text Planning}

Content determination (deciding what information to communicate in the text)
and text-planning (organizing the information into a rhetorically coherent
structure) are done simultaneously in most applied NLG systems
\cite{reiter:arch}.  These tasks can be done at many different levels of
sophistication.  One of the simplest (and most common) approaches is simply
to write a `hard-coded' content/text-planner in a standard programming language
(C++, Lisp, etc).  The resultant system may lack
flexibility, but if the texts being produced have a standardized
content and structure (which is true in many technical
domains), then this can be the most effective way to perform these tasks.

On the other end of the sophistication spectrum, many standard AI techniques
have been adopted for content determination and text planning,
including rule-based systems \cite{idas:aai} and planning \cite{hovy:acl88}.
Systems built in this way are in principal very flexible and powerful,
although in practice they have sometimes not been
robust enough for real-world use.

An intermediate approach which has been quite popular is to use a
special `schema' or `text-planning' language \cite{mckeown:book,kittredge:ci}.
Such languages typically allow the developer to represent text plans as
transition networks of one sort or another, with the nodes giving the
information content and the arcs giving the rhetorical structure
In many cases text-planning languages are implemented as macro packages,
which gives the developer access to the full power of the underlying
programming language whenever necessary.

\section{Sentence Planning}

Sentence planning includes
\begin{itemize}
\item Conjunction and other aggregation.  For example, transforming (1)
into (2):
\begin{itemize}
\item[1)] Sam has high blood pressure.  Sam has low blood sugar.
\item[2)] Sam has high blood pressure {\em and} low blood sugar.
\end{itemize}
\item Pronominalization and other reference.  For example, transforming
(3) into (4):
\begin{itemize}
\item[3)] I just saw Mrs. Black.  Mrs Black has a high temperature.
\item[4)] I just saw Mrs. Black.  {\em She} has a high temperature.
\end{itemize}
\item Introducing discourse markers.  For example, transforming (5) into (6):
\begin{itemize}
\item[5)]If Sam goes to the hospital, he should go to the store.
\item[6)] If Sam goes to the hospital, he should {\em also} go to the store.
\end{itemize}
\end{itemize}
The common theme behind these operations is they do not change the
information content of the text, but they do make it more fluent and easily
readable.

Sentence planning is important if the text needs to read fluently and,
in particular, if it should look like it was written by a human (which is
usually the case for business letters, for example).  If it doesn't matter
if the text sounds stilted and was obviously produced by a computer, then
it may be possible to de-emphasize sentence planning, and perform minimal
aggregation, use no pronouns, etc.

If the text does need to look fluent, then a good job of sentence planning
is essential.  There are formal models of all of the operations mentioned
above, and some applied NLG systems have incorporated them, eg,
\cite{mckeown:anlp94}.  It is also possible to do effective
sentence-planning in an ad-hoc manner, at least in a limited domain;
Knowledge Point's {\em Performance Now} system is a good example of this.

\section{Realization}

A Realizer generates individual sentences (typically from a
`deep syntactic' representation \cite{reiter:arch}).  The realizer needs
to make sure that the rules of English are obeyed, including
\begin{itemize}
\item Point absorption and other punctuation rules.  For example, the sentence
\lingform{I saw Helen Jones, my sister-in-law} should end in ``.'', not ``,.''
\item Morphology.  For example, the plural of \lingform{box} is
\lingform{boxes}, not \lingform{boxs}.
\item Agreement.  For example, \lingform{I am here} instead of
\lingform{I is here}.
\item Reflexives.  For example, \lingform{John saw himself}, instead of
\lingform{John saw John}.
\end{itemize}

There are numerous linguistic formalisms and theories which can be
incorporated into an NLG Realizer, far too many to describe here.
There are also some general-purpose `engines' which can be programmed
with various linguistic rules, such as FUF \cite{elhadad:thesis}
and PENMAN \cite{penman-user-guide}.

In many cases, acceptable performance can be achieved without
using complex linguistic modules.  In particular, if only a few different
types of sentences are being generated, then it may be simpler and cheaper
to use fill-in-the-blank templates for realization, instead of
`proper' syntactic processing.

\section{Conclusion}

Many different techniques are available for performing the three NLG tasks
of content determination (and text planning),
sentence planning, and realization.
These techniques range from the simplistic to the extremely
sophisticated, and it is impossible to say that one is always `better' than
another.  It all depends on the characteristics of the application, such as
whether extensive filtering and summarization of information is needed,
whether texts need to `look like they were written by a person', and
how much syntactic variety is expected to occur in the generated texts.
A good NLG engineer will choose the most appropriate set of
techniques, given the needs of the application \cite{idas:ijcai} and the
available resources.

{\small
\newcommand{\etalchar}[1]{$^{#1}$}

}

\end{document}